\documentclass[conference]{IEEEtran}
\usepackage{amsmath,amssymb,bm}
\usepackage{booktabs,graphicx,microtype,url,xcolor,placeins}
\usepackage[hidelinks]{hyperref}
\graphicspath{{figures/}}

\newcommand{\R}{\mathbb{R}}
\begin{document}
\raggedbottom
\title{CW-WAM: A Structured Contact-Wrench World
Action Model for Force-Sensitive Manipulation}
\author{\IEEEauthorblockN{Houzhu Ding, Xinda Qi, Luqi Wang, Ming Yang}
\IEEEauthorblockA{Ant Group}}
\makeatletter
\newcommand{\teasercaption}[1]{\def\@captype{figure}\caption{#1}}
\makeatother
\IEEEoverridecommandlockouts
\IEEEaftertitletext{%
\begin{minipage}{\textwidth}
\centering
\includegraphics[width=.96\textwidth]{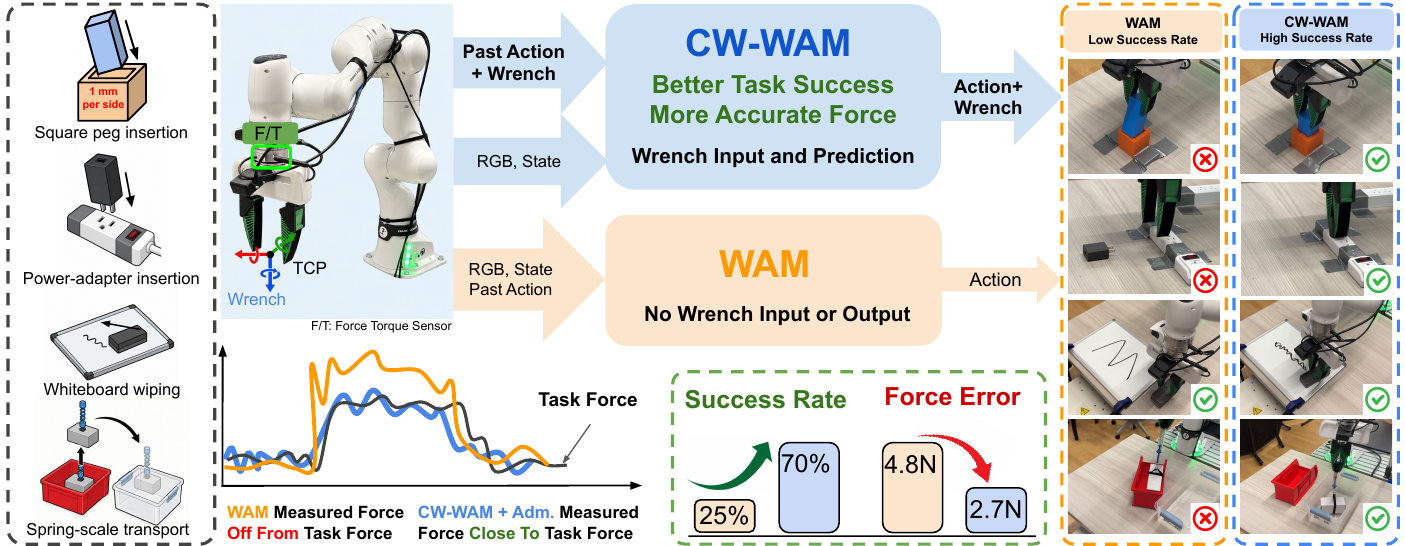}
\setlength{\abovecaptionskip}{2pt}
\teasercaption{CW-WAM helps robots use contact forces to guide their
movements. It learns from past motions and wrist force measurements
to predict both how to move and the forces expected during contact.
These predictions guide motion corrections during execution.
Peg insertion success increases from 25\% with WAM to 70\% with CW-WAM;
adding force feedback to CW-WAM reduces force-tracking error
from 4.8 to 2.7 N.}
\label{fig:teaser}
\end{minipage}\vspace{4pt}}
\maketitle

\begin{abstract}

Force-sensitive manipulation requires contact and load information that is not provided explicitly by 
visual observations.
We present CW-WAM, a world action model that jointly predicts video, motion, and contact
wrench. Inspired by hybrid motion--force control, it contains a unified
interaction module that treats both motion and contact wrench as modeling targets and predicts time-aligned
motion--wrench chunks as control references.
A pair-causal attention mechanism structures the motion--wrench dependencies, while a long fixed-length
wrench memory retains past interaction information. The predicted wrench chunk acts as a
force reference for a 20-Hz outer-loop admittance controller within a slow--fast control system
that maintains force feedback and smooth motion between model queries. CW-WAM improves contact
prediction and increases the insertion success rate from 25\% to 70\% for square pegs with 1-mm clearance per side
and from 40\% to 95\% for power adapters, compared with WAM without wrench input.
It also improves lifting-height accuracy across different loads. 
Our framework with an admittance controller reduces force-reference RMSE
from 4.8 to 2.7 N for peg insertion and from 2.0 to 1.3 N for wiping, respectively,
with smaller force fluctuations during successful execution.
Project website: \href{https://cw-wam-project.github.io/}{\texttt{cw-wam-project.github.io}}.
\end{abstract}

\begin{figure*}[t]
\centering
\includegraphics[width=.96\textwidth,trim=0 0 0 48bp,clip]{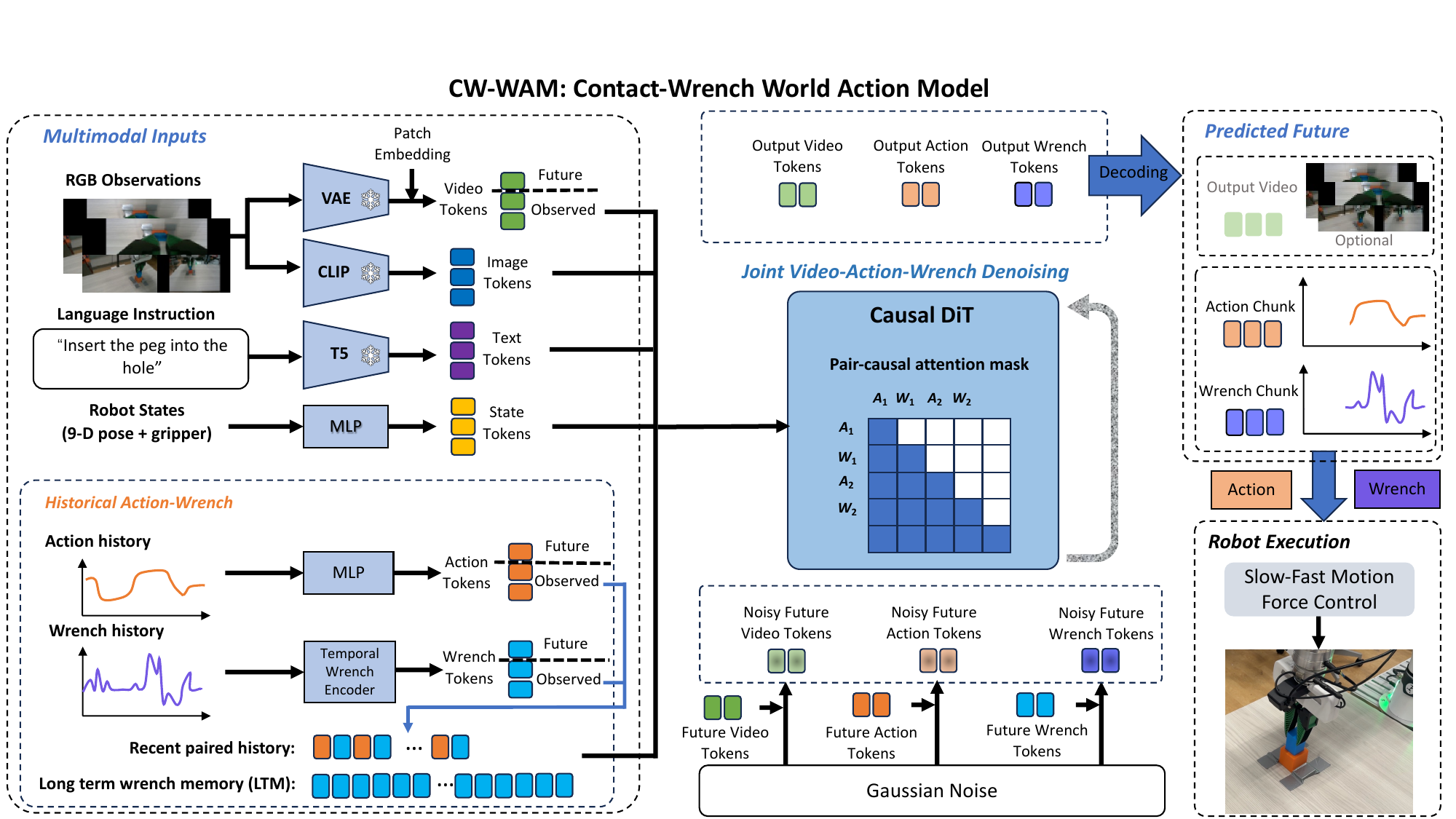}
\setlength{\abovecaptionskip}{2pt}
\caption{CW-WAM jointly predicts video, motion, and wrench from visual observations,
language, robot state, and interaction history. Paired motion--wrench tokens
capture recent interactions, while long-term wrench memory retains earlier
contact information. Pair-causal attention links each future wrench to its paired action
and earlier pairs. The model jointly denoises the three streams using separate prediction heads.
Predicted motion and wrench trajectories provide paired references for robot control.}
\label{fig:architecture}
\end{figure*}

\section{Introduction}
World action models (WAMs) jointly predict future visual observations and actions,
providing coherent motion plans for manipulation~\cite{dreamzero}.
Yet a square peg jammed during insertion may appear fully seated, and
visually similar objects may differ substantially in weight. Such force-sensitive
tasks require physical interaction information that wrist force--torque sensing can provide.

Force feedback improves contact-rich manipulation by complementing visual
observations~\cite{foar,manipforce} and enabling rapid motion correction~\cite{rdp}.
Joint action--wrench prediction extends these benefits: predicted forces guide
learned feedback correction~\cite{fawam} or help evaluate contact
outcomes~\cite{facet0}. These advances motivate WAMs that predict the wrench associated with
each action and use it to guide execution.

Contact depends on commanded motion, prior interaction, and the environment. 
To capture these dependencies, we design an interaction module inspired
by hybrid motion--force control.
The module pairs past commands with measured wrenches
(Fig.~\ref{fig:teaser}) and predicts future motion and wrench
using separate tokens, prediction heads, and training losses.
High-rate wrench encoding captures contact changes within motion intervals, while
longer memory retains earlier wrench measurements.

CW-WAM integrates this module into a DreamZero-style video--action 
backbone~\cite{dreamzero}.
During execution, paired predictions provide nominal motion and 
an expected wrench reference for admittance feedback. 
A slow--fast system applies corrections and tracks continuous motion between model
queries. During contact, this keeps force feedback active while the model
computes the next motion--wrench chunk.
Our work makes three contributions:
\begin{itemize}
\item \textbf{Structured action--wrench modeling.} We introduce a unified
interaction module with paired motion--wrench history, 
separate future output streams and modality-specific losses. 
Across four real-robot tasks, CW-WAM achieves the strongest overall
performance among the evaluated methods. 
It improves insertion and release success, tracks force more accurately
in peg insertion and wiping, and adapts motion to sensed load.

\item \textbf{Predicted wrench references for feedback.} We use the predicted
wrench trajectory to guide admittance corrections without requiring a separately
trained corrective policy. During successful trials, this feedback reduces
the error between predicted and measured forces and lowers contact-force variation.

\item \textbf{A slow--fast execution system.} We combine low-rate CW-WAM inference
with faster admittance feedback and continuous trajectory tracking.
The admittance controller computes pose corrections at 20 Hz. 
The corrected motion waypoints
are then interpolated into smooth trajectories for a 1-kHz robot controller.
We demonstrate this interface on four real-robot tasks.
\end{itemize}

\section{Related Work}
\paragraph{Visuomotor policies and WAMs}

Visuomotor policies generate actions from visual or language-conditioned
observations~\cite{diffusionpolicy,act,pi0,pi05}. 
Alongside direct action generation, world models support policy learning
from imagined trajectories~\cite{daydreamer}.
Video-based approaches recover actions
from generated videos~\cite{unipi,robodreamer,avdc}, while world action
models such as DreamZero jointly model video and actions~\cite{dreamzero}.
Cosmos Policy uses predicted states and values for planning~\cite{cosmospolicy}, 
and Fast-WAM omits future-video generation at inference~\cite{fastwam}.
Building on this foundation, CW-WAM incorporates interaction history 
and wrench prediction.

\paragraph{Contact-aware policies and action--wrench prediction}
Contact-rich manipulation requires interaction signals that vision
does not explicitly provide. 
Prior work incorporates tactile feedback for reactive correction~\cite{rdp}
or visual--tactile--action modeling~\cite{dreamtac,vtwam,vtam}. 
ForceVLA and ManipForce condition actions on F/T
observations~\cite{forcevla,manipforce}, while FoAR modulates force feedback using
predicted contact~\cite{foar}. Gripper-force feedback also supports
learning delicate grasping~\cite{justaddforce}.
Beyond using force as input, FAWAM and Facet-0 jointly predict actions and wrenches. 
They use the predicted wrenches in a learned residual corrector and
an action--wrench critic, respectively~\cite{fawam,facet0}.

CW-WAM pairs past motion commands with measured wrenches and retains
longer wrench histories to capture prior contact.
It predicts future motion and wrench using separate tokens, prediction
heads, and losses. The predicted wrenches then serve as references
for admittance feedback control, without a learned corrective policy.

\paragraph{Slow--fast execution and contact feedback}

Contact can change within an action chunk, requiring a fast response. 
To respond between planning updates, slow planners can guide learned
residual or reflex policies~\cite{phaforce,omnivta}, while learned motion
targets and compliance parameters guide high-rate force 
control~\cite{learningforcecontrol,acp}.
Force Policy combines global visual planning with fast local force 
control~\cite{forcepolicy}.
In CW-WAM, predicted wrench trajectories guide online admittance correction
between model queries.

\section{Structured Contact-Wrench World Action Modeling}
\subsection{Problem formulation and shared backbone}
\label{sec:formulation}
CW-WAM jointly predicts future video, motion, and contact wrench for
motion--force control. At observation time $t$, it models
\begin{equation}
p_\theta(Z_t^+,A_t,W_t
\mid V_t,s_t,\ell,\mathcal H_t^{AW},\mathcal H_t^W).
\end{equation}

Here $V_t$, $s_t$, and $\ell$ denote RGB observations, robot state, and
the language instruction, respectively; $\mathcal H_t^{AW}$ and $\mathcal H_t^W$ provide recent
action--wrench history and longer wrench memory. Outputs are video
latents $Z_t^+$, motion targets $A_t=(a_{t,i})_{i=1}^{H}$, and wrenches
$W_t=(w_{t,i})_{i=1}^{H}$, with hats denoting predictions.
Training pairs each motion target with the compensated wrench at
the same timestamp.

Each action $a_{t,i}\in\R^{10}$ contains position,
continuous 6-D rotation~\cite{rot6d}, and gripper position.

Position and rotation specify a delta from the tool center point (TCP)
pose at observation time $t$, converted to an absolute base-frame pose.

Each predicted wrench $\hat w_{t,i}\in\R^6$ represents
bias-corrected, tool-gravity-compensated force and torque
in the TCP frame defined by its paired predicted pose,
with torque referenced to the TCP origin.

CW-WAM extends DreamZero~\cite{dreamzero} using a Wan2.2-TI2V-5B
DiT~\cite{wan2025}. A VAE encodes three RGB views as a $160\times320$
composite; an MLP encodes robot state. Video, state, history, and
future action--wrench tokens share self-attention, conditioned on
OpenCLIP image and UMT5 language features through cross-attention.

\subsection{Interaction history and paired prediction}

The interaction module pairs past motion commands with measured
wrenches from the same time interval. Recent history spans $H_s$ intervals at rate
$f_h$, with boundaries $b_i=t-(H_s-i)/f_h$. Each interval $(b_{i-1},b_i]$ associates up to
$K_w$ high-rate wrench samples with a motion-target token.

A temporal wrench encoder processes each sample's 6-D value, relative
sampling time, and inter-sample interval. Validity masks exclude padding
from attention and pooling. The action encoder produces a separate token,
and both tokens receive the same history-interval encoding:
\begin{equation}
 \mathcal H_t^{AW}=[A_1^-,W_1^-,\ldots,A_{H_s}^-,W_{H_s}^-].
\end{equation}
Long-term wrench memory (LTM) extends history to $H_\ell$ wrench-only
intervals using the same encoder with additional window-time and history
embeddings. It retains preceding payload tension or sustained resistance.
Both histories overlap and end at the current observation.

Future motion and wrench tokens share the Transformer but use separate
encoders, prediction heads, and losses. Pair-causal (PC) attention uses a
lower-triangular mask over $[A_1^+,W_1^+,\ldots,A_H^+,W_H^+]$.
Thus, $W_i^+$ attends to $A_i^+$, but not conversely; both attend to earlier
pairs. All future tokens update jointly at each denoising step.

\subsection{Training and inference}
\label{sec:training}
Training uses flow matching with normalized action and wrench targets.
For each modality $m\in\{V,A,W\}$, target $x^m$ is mixed with independent
noise $\epsilon^m\sim\mathcal N(0,I)$ at a shared noise
level $\sigma$ within each block:
\begin{align}
x_\sigma^m
&=(1-\sigma)x^m+\sigma\epsilon^m,
\qquad u^m=\epsilon^m-x^m,\\
\mathcal L
&=\sum_{m\in\{V,A,W\}}\lambda_m
\mathbb E\!\left[
\omega(\sigma)
\operatorname{MSE}_{M_m}(v_\theta^m,u^m)
\right].
\end{align}

Here $v_\theta^m$ predicts flow velocity, $\omega(\sigma)$ weights
noise levels, and $M_m$ masks padded targets. Valid-channel normalization
averages motion and wrench losses over their 10 and 6 valid channels,
respectively, with $\lambda_m=1$ for all modalities.
Teacher forcing supplies clean preceding video context to noisy current
blocks~\cite{dreamzero}.

We train the DiT and modality projectors for 50k steps with task-balanced
sampling, keeping VAE, OpenCLIP, and UMT5 frozen. We use $H=H_s=24$,
$H_\ell=150$, $f_h=15$ Hz, and $K_w=16$: 1.6 s of recent history and
10 s of wrench memory.

At inference, four Euler steps jointly generate two video latent
frames and 24 action--wrench pairs spanning 1.6 s, while observed
latents remain fixed. Control uses the paired predictions without
decoding video to RGB.

\begin{figure}[tb]
\centering
\includegraphics[width=.9\columnwidth]{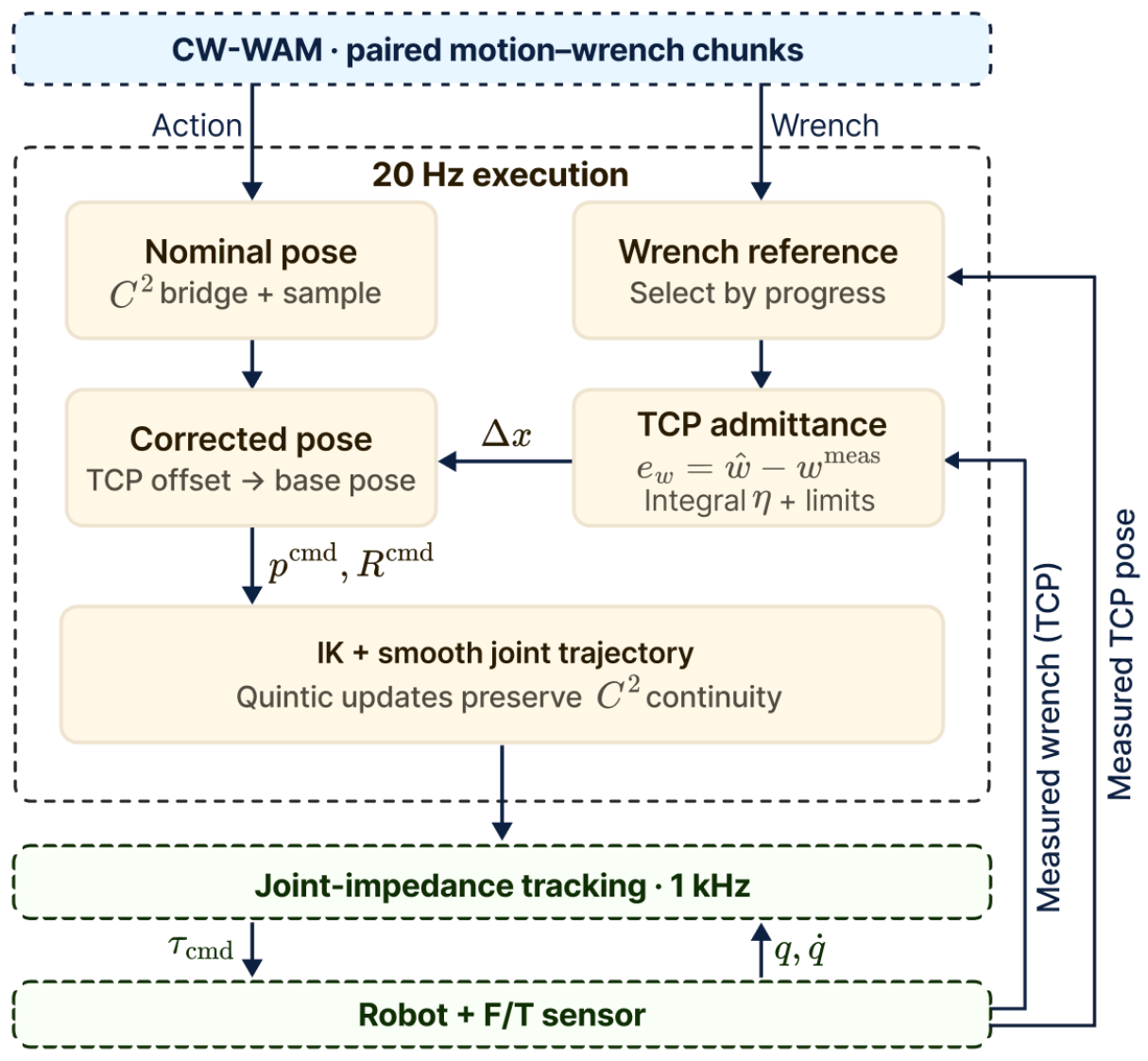}
\caption{Slow--fast motion--force control. Admittance corrects nominal
base-frame poses before IK. Joint-trajectory updates preserve $C^2$
continuity. Wrench feedback is compensated; TCP-pose feedback optionally aligns
wrench progress to measured TCP pose.}

\label{fig:deployment}
\end{figure}

\section{Slow--Fast Motion--Force Control}
\label{sec:control}
We apply task-weighted admittance corrections to nominal motion
at 20 Hz, with joint-impedance tracking at 1 kHz.

\subsection{Admittance control using predicted wrenches}

We use the predicted wrench as a feedback reference to maintain
the interaction expected along the jointly predicted motion.
Predicted motion provides the nominal TCP pose
$(p^{\mathrm{nom}},R^{\mathrm{nom}})$ in the robot base frame.
We express the selected predicted wrench and the measured wrench
in this nominal TCP frame, with moments about its origin,
and compute the admittance correction in the same frame~\cite{modernrobotics,hogan}:

\begin{align}
e_w &= \hat w(\phi^w)-w^{\mathrm{meas}},\\
M\Delta\ddot x+C\Delta\dot x+K_a\Delta x &= -S(e_w+k_I\eta),\\
\dot\eta &= e_w-\lambda\eta.
\label{eq:admittance}
\end{align}
Here $\phi^w$ selects the progress-aligned predicted wrench as described in the next subsection.
$\Delta x=[\Delta p^\top,\Delta\theta^\top]^\top$ contains translation and rotation offsets.
 $M$, $C$, and $K_a$ specify virtual inertia, damping, and restoring stiffness.
The diagonal feedback-weighting matrix $S$ weights the wrench-error components. 
$\eta$ accumulates wrench error with decay rate $\lambda$, and
$k_I$ controls its contribution.
Integral accumulation pauses when correction is inactive or
would increase a saturated offset.

At 20 Hz, semi-implicit Euler integration updates velocity
first and then the offset, subject to acceleration, velocity, and displacement limits.
The corrected base-frame pose is:
\begin{align}
p^{\mathrm{cmd}} &= p^{\mathrm{nom}}+R^{\mathrm{nom}}\Delta p,\\
R^{\mathrm{cmd}} &= R^{\mathrm{nom}}\exp([\Delta\theta]_\times).
\end{align}
Inverse kinematics then maps this corrected absolute pose to joint-position commands.

The feedback-weighting matrix $S$ reflects each task's contact constraints
(Table~\ref{tab:admittance-gains}). For peg insertion, smaller $F_z$ and $\tau_z$ weights
favor lateral and tilt corrections while reducing changes to nominal insertion motion.
Adapter insertion uses larger axial-force and torque weights. Wiping enables only $F_z$
feedback, leaving tangential motion and orientation to the nominal trajectory.

\begin{table}[t]
\centering
\caption{Selected Admittance Gains}
\label{tab:admittance-gains}
\small
\setlength{\tabcolsep}{2pt}
\renewcommand{\arraystretch}{0.95}
\begin{tabular*}{\columnwidth}{@{\extracolsep{\fill}}llrrrrr@{}}
\toprule
Task & Feedback weights & $M$ & $C$ & $K_a$ & $k_I$ & $\lambda$ \\
\midrule
Peg insertion
& $s_T=\operatorname{diag}(1,1,0.05)$ & 5 & 75 & 200 & 0.5 & 0.5 \\
& $s_R=\operatorname{diag}(1,1,0.2)$ & 0.1 & 1.5 & 4 & 0.5 & 0.5 \\
Adapter insertion
& $s_T=\operatorname{diag}(1,1,0.5)$ & 5 & 75 & 200 & 0.1 & 1 \\
& $s_R=\operatorname{diag}(1,1,1)$ & 0.1 & 1.5 & 4 & 0.1 & 1 \\
Whiteboard wiping
& $s_T=\operatorname{diag}(0,0,1)$ & 3 & 34.64 & 100 & 0.1 & 1 \\
& $s_R=\operatorname{diag}(0,0,0)$ & -- & -- & -- & -- & -- \\
\bottomrule
\end{tabular*}
\par\smallskip
{\footnotesize $S=\operatorname{blkdiag}(s_T,s_R)$; $s_T$ and $s_R$ weight translational and rotational feedback, respectively. $k_I$ and $\lambda$ are in $\mathrm{s}^{-1}$.\par}
\end{table}

\subsection{Continuous action-chunk execution}

On a single NVIDIA RTX 5090, four-step model queries typically take
0.5--0.6 s, including observation processing, inference, and postprocessing,
but excluding network communication. Chunks execute sequentially
(Fig.~\ref{fig:deployment}). While awaiting the next prediction, the robot
holds the final nominal pose with admittance active during contact.
An adaptive-duration $C^2$ bridge joins each new chunk to the current
commanded pose, matching velocity and acceleration.

At 20 Hz, we interpolate nominal poses with $C^2$ continuity and
predicted wrenches linearly. All experiments select the wrench reference
by matching the measured TCP pose to the planned motion trajectory.
Matched progress $\phi^w$ cannot move backward or exceed scheduled
progress $\phi$, accommodating execution lag while preserving reference order.

Every 50 ms, we replan a 0.5-s piecewise-quintic joint trajectory,
matching the active trajectory's position, velocity, and acceleration for
$C^2$ continuity. A 1-kHz joint-impedance controller tracks it with gravity
and friction compensation.

\section{Experiments}

We evaluate prediction accuracy through open-loop ablations,
force-sensitive manipulation through real-robot tasks, and execution
quality with and without predicted-wrench admittance.

\subsection{Tasks and data collection}

We select four tasks to test contact-guided alignment, precise seating, force regulation,
and adaptation to payload (Fig.~\ref{fig:teaser}, left).

\textbf{Square peg insertion.} The robot inserts a 38-mm square peg into a 40-mm hole
with 1-mm clearance per side and an insertion depth of about 50 mm. The tight clearance and
long engagement make misalignment and jamming challenging. Demonstrations include smooth
first-attempt insertion, failed attempts followed by recovery, and light
jamming followed by alignment. Insertion success requires full seating;
release success additionally requires the peg to remain seated after gripper opening.

\textbf{Power-adapter insertion.} The robot grasps a loosely placed adapter and seats it
in another socket. This tests precise positioning and insertion with contact.
Compared with the long peg, the adapter is less prone to jamming once insertion begins.
Demonstrations therefore emphasize successful first-attempt insertion with minimal lateral
adjustment. The success criteria are the same as for peg insertion.

\textbf{Whiteboard wiping.} The robot picks up an eraser and removes a marker trace
through sustained surface contact. The traces vary in orientation and curvature,
requiring both spatial coverage and sustained force regulation. Success requires complete marker
removal after automatic eraser pickup. We also quantify ink removal in additional
patterned-wiping recordings and evaluate contact-force accuracy separately.

\textbf{Spring-scale transport.} The robot grasps a spring scale attached to a box
and transfers the box between bins. The boxes weigh either 30 g or 830 g but look identical. The heavier box stretches
the spring farther, requiring a higher lift to clear the rim. This tests whether contact
and load information guide motion beyond visual appearance. Success requires rim clearance,
transport, and placement in the destination bin.

Training uses task-specific language instructions and 450 demonstrations:
213 peg insertions, 97 adapter insertions, 64 whiteboard wipes,
and 76 spring-scale transports.

Data are collected by bilateral teleoperation with two 7-DoF Franka arms.
Leader joint positions command the follower, whose external-joint-torque estimates
provide force feedback to the operator. A wrist-mounted LinkTouch six-axis F/T sensor
records wrench at 150 Hz. We remove sensor bias and calibrated tool gravity, transform
the wrench to the TCP frame with moment about the TCP origin, and apply a causal
second-order Butterworth low-pass filter with a 10-Hz cutoff. The resulting wrench
represents external contact and grasped-payload loads. Two RealSense D405 cameras and
one D455 provide three RGB views. Motion commands, robot state, and images are recorded
at 15 Hz and aligned with the wrench stream by timestamps.

Training uses eight NVIDIA H20 GPUs and AdamW with a global
batch size of 8, a peak learning rate of $10^{-5}$, and weight
decay of $10^{-5}$. We use 5\% linear warmup followed by cosine decay.

\subsection{Model variants}
\label{sec:model-variants}
\textbf{WAM} has no wrench input or output. \textbf{WAM + Wrench} encodes pooled
wrench history and predicts concatenated 16-D action--wrench vectors.
It uses the current robot state without explicit past motion-command tokens.
\textbf{CW-WAM w/o LTM} introduces the interaction module: explicit paired command
and wrench history, separate future streams, and PC attention.
Its recent history pairs 24 motion commands with wrench intervals over 1.6 s.
Full \textbf{CW-WAM} adds LTM.
\textbf{CW-WAM w/o LTM and PC} replaces only PC with bidirectional
attention, isolating attention structure within the no-LTM module.

All variants share training data, backbone initialization, and a 50k-step
schedule with task-balanced sampling. WAM omits wrench loss.
WAM + Wrench uses either padded-width normalization, dividing masked
squared-error sums by 32, or CW-WAM's valid-channel normalization
(Sec.~\ref{sec:training}), with architecture and other settings unchanged.
The latter tests the interaction module under the same loss normalization.
PC and LTM ablations retain the CW-WAM objective: PC is tested without
LTM, and LTM with PC retained.

\begin{figure*}[t]
\centering
\includegraphics[width=.94\textwidth]{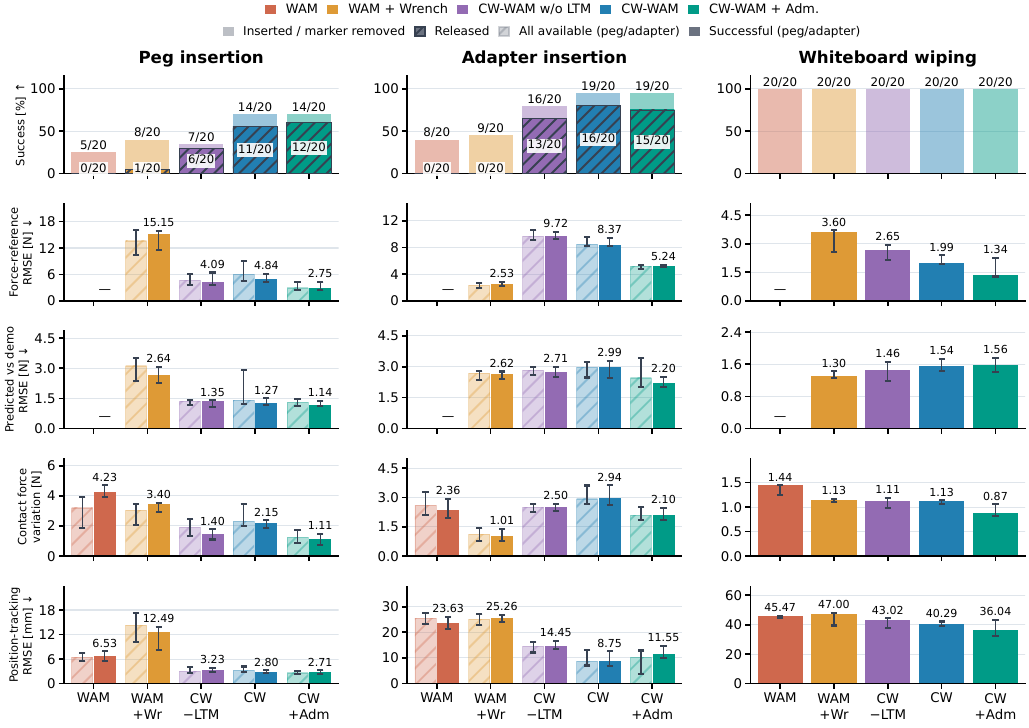}
\caption{Task outcomes and execution quality. Top: insertion or marker-removal
success out of 20 trials; hatched bars show successful release. Quality rows
compare all 20 trials (light, hatched) with successful subsets
(dark bars); these groups coincide for wiping. A dash indicates no wrench prediction.}
\label{fig:task-summary}
\end{figure*}

\subsection{Contact-phase prediction and ablations}
\label{sec:open-loop}
\begin{table}[!t]
\centering
\caption{Open-loop prediction across four tasks.
Padded and valid denote padded-width and valid-channel loss normalization.
Bold: lowest error; --: no wrench output}
\label{tab:dense-main}
\small
\footnotesize
\setlength{\tabcolsep}{1.5pt}
\begin{tabular*}{\columnwidth}{@{\extracolsep{\fill}}lrrrr@{}}
\toprule
Model &
\shortstack{$e_p$\\{[mm]}} &
\shortstack{$e_R$\\{[$^\circ$]}} &
\shortstack{$e_F$\\{[N]}} &
\shortstack{$e_I$\\{[N\,s]}} \\
\midrule
WAM & 26.97 & 2.85 & -- & -- \\
WAM + Wrench (padded) & 26.32 & 2.76 & 3.15 & 2.33 \\
WAM + Wrench (valid) & 25.50 & 2.80 & 3.03 & 2.21 \\
CW-WAM w/o LTM and PC & 22.88 & 2.25 & 2.92 & 2.11 \\
CW-WAM w/o LTM & \textbf{19.92} & \textbf{2.02} & 2.84 & 2.17 \\
CW-WAM & 20.69 & 2.07 & \textbf{2.79} & \textbf{2.01} \\
\bottomrule
\end{tabular*}
\end{table}

Table~\ref{tab:dense-main} evaluates 50k-step checkpoints on 373
24-step windows from 22 test episodes, using four denoising steps and seed 42.
Windows use stride 24 plus a final tail window. Contact means
$\|F\|_2\geq5$ N; dynamic windows contain contact and satisfy
$\max_{i,j}\|F_i-F_j\|_2\geq2$ N. We average windows within episodes, 
then episodes within tasks, weighting tasks equally.

Contact-position error $e_p$ is the mean Euclidean distance, 
and contact-orientation error \(e_R\) is the mean SO(3) geodesic angle between
predicted and recorded motion targets at contact samples.
Dynamic force-vector RMSE $e_F$ measures
$\sqrt{\operatorname{mean}_i
\|\hat{F}_i-F_i^{\mathrm{meas}}\|_2^2}$
at recorded timestamps, without temporal shifting.
All models use the same 20-step interior of each selected window,
retaining approach, loading, and release samples.
Force-magnitude integral error $e_I$ is the absolute difference
between the trapezoidal integrals of predicted and measured force
magnitudes over all 24 steps, complementing pointwise RMSE with
cumulative force-magnitude agreement.

Valid-channel normalization lowers WAM + Wrench errors from 26.32 to
25.50 mm and from 3.15 to 3.03 N.
Under the same normalization,
CW-WAM without LTM achieves 19.92 mm and 2.84 N: paired reductions
of 5.58 mm (95\% CI [3.65, 7.84]) and 0.19 N (95\% CI [0.04, 0.33]).
Orientation error also decreases 
from 2.80$^\circ$ to 2.02$^\circ$, and dynamic torque-vector RMSE from 0.0908 to 0.0837 N\,m.
These gains support structured interaction modeling beyond loss normalization.
CIs use 10,000 paired episode bootstrap resamples, stratified
by task, with trained models held fixed.

PC reduces no-LTM contact-position and orientation errors
by 2.96 mm (95\% CI [2.05, 3.89]) and
0.23$^\circ$ (95\% CI [0.12, 0.35]), respectively. Full CW-WAM achieves the lowest dynamic
force-vector RMSE (2.79 N) and force-magnitude integral error (2.01 N\,s).
The no-LTM variant retains slightly lower position and orientation errors.
Real-robot tests evaluate how these gains translate into interaction
and task completion.

\subsection{Real-robot evaluation and metrics}

Real-robot tests evaluate WAM, WAM + Wrench (padded), CW-WAM w/o LTM,
and CW-WAM, then add admittance to full CW-WAM.
Loss normalization and PC are compared in open-loop tests; real-robot
tests focus on the interaction module, LTM, and admittance feedback. 

All tests use 50k-step checkpoints. Before each trial, we reset the arm and
gripper and randomly place objects near their demonstration locations,
within the demonstration distribution.

For peg insertion and wiping, we compare force accuracy and variation
under demonstration-median and CW-WAM-predicted force references.
The demonstration reference is the componentwise median of phase-aligned 
training force profiles.

An operator verifies outcomes from video using the task criteria.
Each method uses 20 trials per contact task and 10 per load
(Sec.~\ref{sec:load-lifting}). Fig.~\ref{fig:task-summary} reports
success rates and quality metrics for all 20 trials and successful
subsets, which coincide for wiping.

Across real-robot analyses, continuous-valued metrics are
summarized by medians and interquartile ranges (IQRs).
Aggregate curves show pointwise medians with shaded IQR bands.

Position-tracking RMSE compares commanded and measured TCP positions;
force-reference RMSE is $\sqrt{\operatorname{mean}\|F^{\mathrm{ref}}-F^{\mathrm{meas}}\|_2^2}$,
where $F^{\mathrm{ref}}$ is the selected force reference.
Predicted--demonstration RMSE compares predicted force magnitudes with the
phase-aligned demonstration median (Fig.~\ref{fig:task-summary}).
Force and torque variations are the duration-weighted means
of local standard deviations of their respective magnitudes,
using centered 5-s windows over contact periods.

Fig.~\ref{fig:contact-force-quality} compares measured and demonstration
force magnitudes over contact phases identified from video and force,
with low-force intervals compressed.
Fig.~\ref{fig:payload-response} shows phase-aligned lifting heights.
For the feedback comparison (Fig.~\ref{fig:force-tracking}), contact intervals
are detected from measured force, with gripper state and robot motion
used to exclude pickup and free motion.

\subsection{Task completion and contact behavior}

The interaction module improves insertion and release (Fig.~\ref{fig:task-summary}). Relative
to WAM + Wrench, CW-WAM w/o LTM raises adapter insertion from 9/20 to 16/20 and release from
0/20 to 13/20. Peg release improves from 1/20 to 6/20, with insertion at 7/20 versus 8/20.

\begin{figure}[tb]
\centering
\includegraphics[width=.95\linewidth]{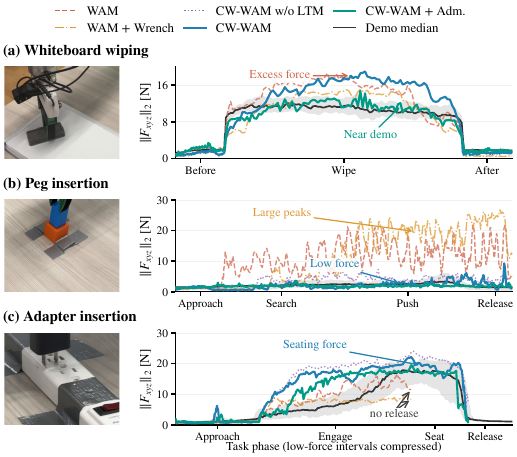}
\caption{Contact-force curves across tasks. Colored curves show measured
force magnitudes across successful trials. Thick blue and green curves show CW-WAM
without and with admittance, respectively. The black curve and gray band show the
demonstration reference. Curves are aligned by contact phase, with low-force
intervals compressed.
Adapter unloading includes only trials with successful release.}
\label{fig:contact-force-quality}
\end{figure}

Longer wrench memory improves both insertion tasks: peg insertion rises from 7/20 to 14/20 and
release from 6/20 to 11/20; adapter insertion rises from 16/20 to 19/20 and release from 13/20
to 16/20. It also lowers force-reference error in adapter insertion and wiping, demonstrating
the task value of longer interaction context (Fig.~\ref{fig:task-summary}).

In successful peg trials, CW-WAM reduces position-tracking RMSE from 12.5 to 2.8 mm and
force-reference RMSE from 15.2 to 4.8 N relative to WAM + Wrench (Fig.~\ref{fig:task-summary}).
Force profiles also show less variation during search and seating
(Fig.~\ref{fig:contact-force-quality}(b)).

For adapter insertion, CW-WAM improves completion and reduces position-tracking RMSE from 25.3 to 8.8 mm. 
Its higher force-reference RMSE and contact-force variation relative to WAM + Wrench (Figs.~\ref{fig:task-summary} and \ref{fig:contact-force-quality}(c))
are consistent with the observation that most successful WAM + Wrench trials rely on precise, one-shot alignment without
adjustment, followed by full insertion. In contrast, CW-WAM’s learned search and firm insertion
behavior improves insertion success while producing higher contact forces.

In the whiteboard-wiping task, all methods remove the ink in 20/20 trials with simple patterns. For more complicated
patterns, such as circles and letters, the average removal rate is 85.9\% across the six selected trace patterns.
For contact-force quality, force-reference RMSE and contact-force variation both decrease from WAM + Wrench to CW-WAM with admittance, 
showing more accurate force tracking and smoother contact (Figs.~\ref{fig:task-summary} and \ref{fig:contact-force-quality}(a)).

\subsection{Load-dependent lifting}
\label{sec:load-lifting}

CW-WAM reproduces load-dependent lifting heights more accurately than the other models,
even when the boxes look identical.

\begin{figure}[t]
\centering
\includegraphics[width=.95\linewidth]{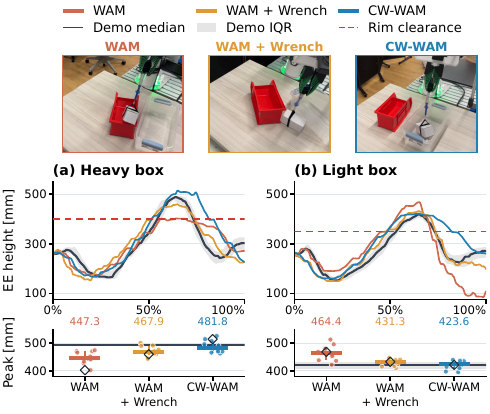}
\caption{Load-dependent lifting under light and heavy payloads.
Curves show representative end-effector height trajectories. 
Dashed lines indicate approximate heights needed
to clear the bin rim.
The lower panels show
peak heights from all ten trials per method and load.
Diamonds mark the trials shown in the trajectories. Gray lines and bands show
demonstration references.}
\label{fig:payload-response}
\end{figure}

Its median peak heights are 423.6 and 481.8 mm for light and
heavy loads, respectively,
close to the demonstration medians of 420.9 and 492.7 mm (Fig.~\ref{fig:payload-response}). CW-WAM w/o LTM shows similar lifting behavior
(not plotted). WAM reaches 464.4 and 447.3 mm, showing no higher
lift for heavier loads. WAM + Wrench gives an intermediate response at 431.3 and
467.9 mm: its absolute differences from the demonstration medians are 10.4 and
24.8 mm, compared with 2.7 and 10.9 mm for CW-WAM.

\subsection{Predicted-wrench feedback}

Admittance further reduces force-reference RMSE and measured-force
variation compared with CW-WAM alone (Fig.~\ref{fig:task-summary}).
Peg RMSE decreases from 4.8 to 2.7 N and wiping RMSE from 2.0 to
1.3 N, while insertion and wiping success rates remain unchanged.

\begin{figure}[tb]
\centering
\includegraphics[width=.95\linewidth]{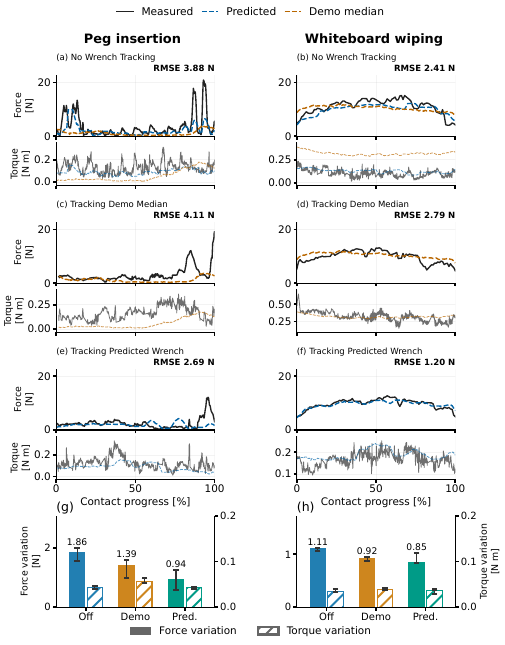}
\caption{Force and torque tracking for peg insertion (left) and
whiteboard wiping (right). (a--f) Measured and reference magnitudes,
with three-axis force RMSE labels. Each example is selected from the trial whose 
force variation is closest to the group median. (g,h) Force variation (solid bars, left axis) and
torque variation (hatched bars, right axis). Aggregate bars summarize 10 trials per condition.}
\label{fig:force-tracking}
\label{fig:fixed-reference-tracking}
\end{figure}

Fig.~\ref{fig:force-tracking} compares no wrench tracking,
demonstration-median wrench tracking, and predicted-wrench tracking.
For the representative trials in Fig.~\ref{fig:force-tracking},
force-reference RMSEs are 3.88 and 2.41 N without feedback
(a, b), and 2.69 and 1.20 N with predicted-wrench tracking (e, f).

Predicted-wrench tracking also yields the lowest median contact-force
variation, which decreases from 1.86 to 1.39 to 0.94 N for peg
insertion and from 1.11 to 0.92 to 0.85 N for wiping. The larger
reduction in peg insertion, where contact changes during search
and seating, highlights the value of predicted wrench references
for smoother contact.
Torque errors and variations are generally small across tasks. 

\section{Conclusion and Discussion}

In this work, we presented CW-WAM, a world action model that jointly
predicts video, motion, and contact wrench. The interaction module
combines paired action--wrench history, separate prediction streams,
and longer wrench memory. Across four real-robot tasks, CW-WAM
improves insertion and release success rates and adapts the lifting
height to different loads. Compared with the WAM baseline without
wrench input, the peg insertion success rate increases from 25\%
to 70\%, and the adapter insertion success rate increases from
40\% to 95\%.

The slow--fast execution system uses predicted wrenches as references
for admittance feedback control between model queries. During
successful execution, this feedback mechanism reduces force-reference RMSE
from 4.8 N to 2.7 N for peg insertion and from 2.0 N to 1.3 N for
wiping, while also reducing contact-force variation.

Our evaluation covers four tasks within the demonstration
distribution, using manually tuned feedback gains and predefined
compliant directions. Future work will examine how controller
settings affect performance and explore learning additional contact
parameters such as the feedback-weighting matrix and contact frame,
alongside motion and wrench prediction for end-to-end
closed-loop control. We will also evaluate generalization to new
objects and tasks.

\section*{Acknowledgments}
Under the authors' guidance, OpenAI Codex assisted with polishing
the Abstract and Sections I--VI, code development, and plotting scripts 
supporting Figs.~1, 4--7.

\end{document}